\documentclass{article}
\usepackage{microtype}
\usepackage{graphicx}
\usepackage{subcaption}
\usepackage{booktabs}
\usepackage{multirow}
\usepackage{hyperref}

\usepackage[preprint]{icml2026}

\usepackage{amsmath}
\usepackage{amssymb}
\usepackage{mathtools}
\usepackage{amsthm}
\usepackage[capitalize,noabbrev]{cleveref}

\theoremstyle{plain}

\theoremstyle{definition}

\theoremstyle{remark}

\icmltitlerunning{Focus on What Matters}

\begin{document}

\twocolumn[
\icmltitle{Focus on What Matters: Two-Stage ROI-Aware Refinement for Anatomy-Preserving Fetal Ultrasound Reconstruction}

\begin{icmlauthorlist}
\icmlauthor{Ines Abbes}{hbku}
\icmlauthor{Mahmood Alzubaidi}{hbku}
\icmlauthor{Mowafa Househ}{hbku}
\icmlauthor{Khalid Alyafei}{sidra}
\icmlauthor{Marco Agus}{hbku}
\icmlauthor{Samir Brahim Belhaouari}{hbku}
\end{icmlauthorlist}

\icmlaffiliation{hbku}{College of Science and Engineering, Hamad Bin Khalifa University, Doha, Qatar}
\icmlaffiliation{sidra}{Sidra Medicine, Doha, Qatar}

\icmlcorrespondingauthor{Ines Abbes}{inab11975@hbku.edu.qa}

\icmlkeywords{medical image reconstruction, fetal ultrasound, ROI-aware learning, domain shift, representation learning}

\vskip 0.3in
]

\printAffiliationsAndNotice{}

\begin{abstract}
Measurement-critical ultrasound tasks often depend on a small anatomical region, making global reconstruction metrics an unreliable proxy for clinical fidelity. We propose an ROI-aware representation learning framework and instantiate it for first-trimester nuchal translucency (NT) screening under multi-hospital domain shift. A two-phase convolutional autoencoder (CAE) first learns a globally faithful 128-D latent code via MS-SSIM, then refines the NT ROI using intensity ($L_1$) and normalized Sobel-edge constraints. To combine these heterogeneous objectives without manual tuning, we initialize loss weights via gradient-based calibration from per-term gradient magnitudes. Under strict hospital-wise evaluation with one hospital held out, ROI refinement improves both global and measurement-relevant quality: on the standard dev split it increases PSNR by +0.27\,dB (val) and +0.29\,dB (held-out test), reduces ROI MAE by 8.87\% (val) and 6.43\% (held-out test), and reduces ROI Edge-MAE by 11.10\% on source hospitals and 4.90\% on the unseen hospital. Beyond reconstruction, frozen-latent probes provide additional evidence of generalization: hospital provenance becomes less confidently predictable on the unseen site (0.556$\rightarrow$0.541 max-softmax; 0.684$\rightarrow$0.688 entropy) while OOD detection remains strong across site-held-out protocols (Mahalanobis AUROC up to 0.9956, with modest KNN gains in challenging splits). The same ROI-aware refinement principle is anatomy-agnostic and can be adopted for other fetal biometry targets (e.g., crown--rump length (CRL), nasal bone (NB)) and broader medical imaging settings where small ROIs dominate clinical decisions.
\end{abstract}

\section{Introduction}
\label{sec:introduction}

First-trimester prenatal screening constitutes a cornerstone of modern obstetric care, enabling early detection of chromosomal abnormalities and structural anomalies that inform critical clinical decisions~\cite{kasera2024,li2024nt}. Central to this screening protocol is the measurement of the Nuchal Translucency (NT) ---a fluid-filled space at the posterior aspect of the fetal neck whose thickness between 11--14 weeks of gestation provides essential prognostic information~\cite{hur2024}. Accurate NT measurement has been validated across multiple large-scale studies as a reliable marker for Down syndrome and other aneuploidies~\cite{dalton2005}, with standardized protocols established by international guidelines~\cite{salomon2013}. Beyond chromosomal screening, increased NT thickness serves as a predictor for congenital heart defects and other structural anomalies~\cite{zhang2024nt}, underscoring the clinical imperative for precise and reproducible NT visualization.

Despite its clinical importance, obtaining consistent NT measurements remains challenging due to the inherent characteristics of ultrasound imaging. Ultrasound images are formed through acoustic wave propagation and echo reflections, producing \emph{speckle}---a multiplicative, texture-like noise pattern that can obscure fine anatomical boundaries~\cite{hu2024}. This challenge is compounded by acoustic artifacts including shadowing from attenuating structures, signal dropout at unfavorable beam angles, and reverberation patterns that introduce spurious bright regions. Furthermore, the same anatomical structure can appear markedly different across imaging systems due to scanner-dependent post-processing pipelines that apply varying levels of gain compensation, dynamic range compression, and edge enhancement~\cite{boumeridja2025}.

Operator variability introduces an additional layer of inconsistency. Studies have documented substantial intra- and inter-observer variability in fetal biometric measurements~\cite{ramesh2025}, arising from differences in probe positioning, applied pressure, and plane selection. The NT region is particularly sensitive to these factors: even slight deviations from the mid-sagittal plane can alter the apparent thickness of this narrow fluid-filled space. In multi-center clinical settings, this variability is amplified by differences in equipment, acquisition protocols, and patient populations across sites---creating a domain shift problem that undermines the generalizability of automated methods.

\paragraph{Brief context and limitation of existing approaches.}
Recent deep learning methods have advanced medical image reconstruction and enhancement, including diffusion probabilistic models~\cite{kazerouni2024} and transformer-based designs such as Swin Transformers~\cite{tang2022} and hybrid CNN-Transformer architectures~\cite{chen2025medfusenet}. For ultrasound, GAN-based enhancement and super-resolution approaches have also shown promising perceptual improvements across devices and acquisition settings~\cite{athreya2024cyclegan,boumeridja2025}. However, most of these methods are optimized primarily with \emph{global} reconstruction objectives, implicitly assuming that improving whole-image fidelity guarantees clinically meaningful preservation. In NT screening, the clinically relevant region occupies only a small fraction of the image, and global losses can be dominated by background appearance while allowing subtle but consequential changes in NT anatomy.

\paragraph{Problem statement.}
We address the problem of \emph{measurement-relevant} reconstruction/representation learning for first-trimester NT ultrasound: learning a compact representation that reconstructs the image while explicitly preserving NT intensity and boundary fidelity inside a small Region Of Interest (ROI) under cross-hospital domain shift. We observe three recurring failure modes under global-only optimization: (i) \emph{boundary softening} that blurs caliper-relevant edges, (ii) \emph{boundary shifts} of a few pixels that bias measurement, and (iii) \emph{structure attenuation} where thin membrane cues become faint or disappear. These limitations motivate an ROI-aware learning strategy that constrains NT appearance directly while maintaining globally coherent structure, with the goal of improving robustness and transfer to unseen clinical sites.

\paragraph{Contributions.}
We propose a two-phase Convolutional Autoencoder (CAE) framework that learns a compact latent representation optimized for anatomy preservation under domain shift. Our contributions are as follows:
\begin{itemize}
    \item \textbf{ROI-aware two-phase representation learning:} We introduce a phased training strategy where Phase~1 learns globally faithful features using a multi-scale structural similarity objective, establishing a stable latent anchor. Phase~2 then refines the representation using explicit intensity and Sobel-based edge constraints within the NT ROI, forcing the latent code to encode measurement-critical boundaries without sacrificing global consistency. Although ROI-aware masking has recently been explored for masked autoencoders in medical ultrasound videos~\cite{szijarto2024masked}, to the best of our knowledge ROI-based refinement formulations that explicitly target measurement-critical landmark fidelity in reconstruction have not yet been exploited in the medical field. The formulation is ROI-agnostic and can be adopted for other fetal biometry targets (e.g., CRL, NB) and broader medical imaging settings.

    \item \textbf{Gradient-based multi-objective loss calibration:} To address the challenge of combining objectives with disparate numeric scales, we propose an initialization strategy that calibrates loss weights based on per-term gradient magnitudes. This yields comparable influence across objectives at training onset without manual hyperparameter tuning, providing a reproducible alternative to trial-and-error weighting that is particularly beneficial under cross-hospital variability.

    \item \textbf{Hospital-wise generalization with downstream validation of the representation:} We evaluate under a strict site-disjoint protocol where one hospital is held out entirely for testing. Beyond reconstruction metrics, we validate the learned (frozen) encoder representation on multiple downstream tasks described in Section~\ref{sec:downstream_framework}, including provenance classification, Quality Control (QC) triage, and Out Of Distribution (OOD) detection, to provide additional evidence that the representation transfers across hospitals. Overall, the proposed refinement reduces NT boundary error by 11\% on source sites and 5\% on the held-out target site.
\end{itemize}

A detailed discussion of the related literature and our positioning relative to the state-of-the-art can be found in Appendix~\ref{app:literature} to keep the main narrative focused.
\section{Method}
\label{sec:method}
\begin{figure*}[t]
  \centering
  \includegraphics[width=\textwidth]{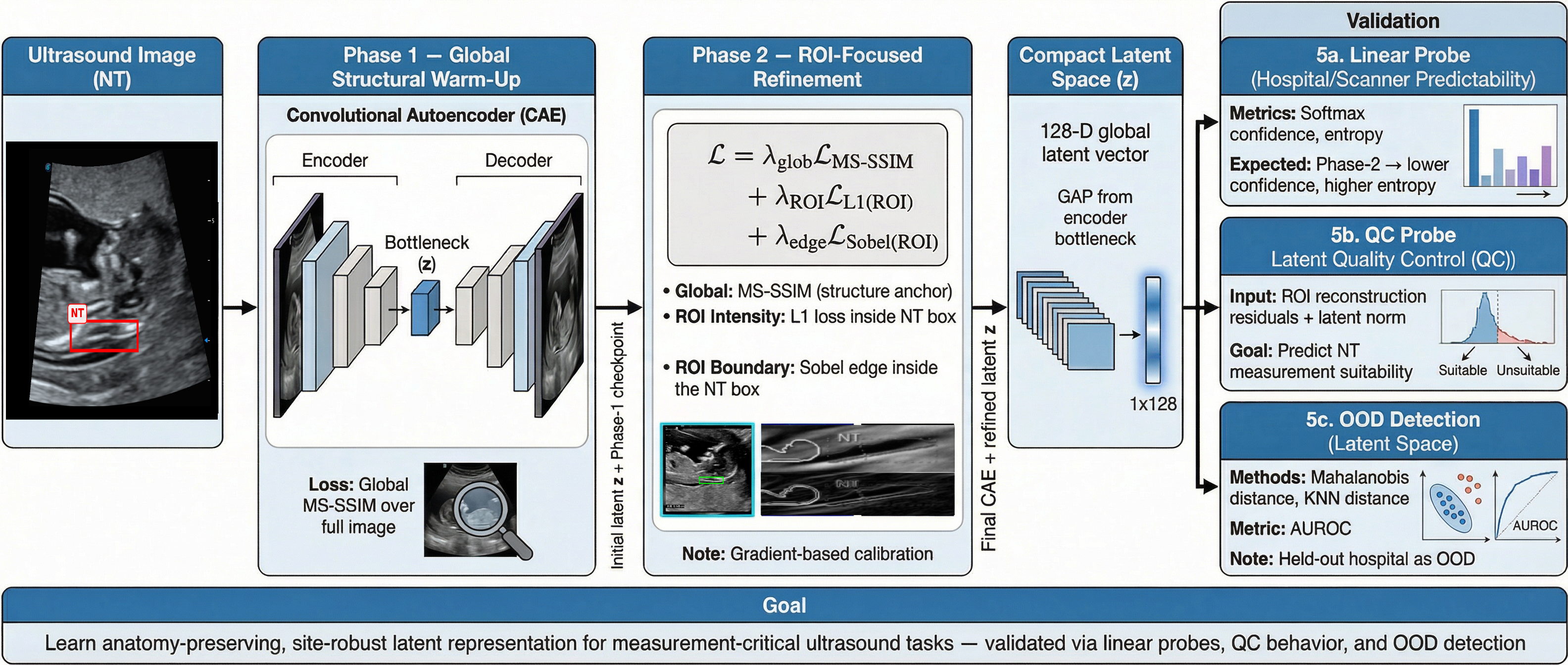}
  \caption{Overview of the Two-Phase Training Strategy. \textbf{Phase 1} warms up the encoder with global structural similarity. \textbf{Phase 2} freezes the global structure and refines the ROI using localized intensity (L1) and gradient (Sobel) losses, balanced via gradient calibration.}
  \label{fig:two_phase_overview}
\end{figure*}

Our ROI-aware representation learning framework for fetal ultrasound is depicted in Fig.~\ref{fig:two_phase_overview}. Unlike standard reconstruction or pure detection approaches (e.g., YOLO~\cite{wang2024yolov9}), our goal is to learn a compact latent representation that preserves measurement-critical anatomy (NT) while capturing broader domain characteristics. This representation serves as a versatile foundation for downstream tasks including measurement, quality control, and site provenance analysis.

\subsection{Problem Formulation and Design Rationale}
\label{sec:problem}

Let $x \in \mathbb{R}^{H \times W}$ be a grayscale ultrasound image. In first-trimester screening~\cite{snijders1998uk}, clinical utility depends on a specific anatomical ROI, denoted by the pixel set $\Omega \subset \{1,\dots,H\} \times \{1,\dots,W\}$.
Standard autoencoders optimize a global similarity metric (e.g., MSE or SSIM~\cite{wang2003multiscale}) over the full image. We observe that this leads to three ROI-specific failure modes in the NT setting:
(i) \textbf{Boundary softening}, where critical caliper placement edges become blurred;
(ii) \textbf{Spatial shifts}, where interfaces migrate slightly, corrupting measurement; and
(iii) \textbf{Structure loss}, where thin membrane structures disappear entirely.

Our objective is to learn a mapping to a low-dimensional latent space $z$ such that the reconstruction $\hat{x}$ retains high fidelity specifically within $\Omega$, without sacrificing global structural coherence.

\subsection{Dataset and Preprocessing}
\label{sec:data-preproc}

\textbf{Multi-center Data.} 
We utilize a multi-center ultrasound dataset collected from three hospitals (Hospital-1, Hospital-2, Hospital-3) to explicitly evaluate robustness to real-world domain shifts, as illustrated in Figure~\ref{fig:hospital_samples}. 
The dataset comprises a total of 2{,}028 fetal ultrasound images acquired during the first trimester of pregnancy, including 766 images from Hospital-1, 759 images from Hospital-2, and 503 images from Hospital-3. 
All images correspond to standard NT screening scans performed between approximately 11 and 13+6 weeks of gestation, when NT measurement is clinically recommended.
\begin{figure}[t]
  \centering
  \includegraphics[width=0.95\columnwidth]{NT_examples_three_hospitals_set1.png}
  \caption{Representative samples from the three clinical sites showing the NT region (red box). Note the significant domain shift in contrast, speckle texture, and field-of-view across hospitals (Hosp-1, Hosp-2, Hosp-3).}
  \label{fig:hospital_samples}
\end{figure}
All images are provided in the standard NT view and are accompanied by expert-annotated NT bounding boxes, enabling both ROI-aware reconstruction and region-specific evaluation.
To assess cross-site generalization, we adopt a \emph{site-held-out (leave-one-site-out)} evaluation protocol, in which models are trained and validated on data from two hospitals and evaluated on the remaining unseen hospital, with 15\% of the training data reserved for validation in each configuration (see Table~\ref{tab:dataset_dist}). 
Due to privacy restrictions, the dataset is not publicly available but can be shared upon request for research purposes.

\begin{table}[t]
\centering
\caption{Leave-one-site-out evaluation protocol with explicit train/validation/test sizes. 
For each split, two hospitals are used for training and validation (15\% held out for validation), while the remaining hospital is used exclusively for testing.}
\label{tab:dataset_dist}
\vskip 0.05in
\resizebox{\columnwidth}{!}{%
\begin{tabular}{lcccc}
\toprule
\textbf{Split} & \textbf{Train Hospitals} & \textbf{\# Train} & \textbf{\# Val} & \textbf{\# Test} \\
\midrule
H1 + H2 $\rightarrow$ H3 & H1, H2 & 1296 & 229 & 503 \\
H1 + H3 $\rightarrow$ H2 & H1, H3 & 1079 & 190 & 759 \\
H2 + H3 $\rightarrow$ H1 & H2, H3 & 1073 & 189 & 766 \\
\bottomrule
\end{tabular}%
}
\vskip -0.1in
\end{table}

\textbf{Input Standardization.} Ultrasound images vary significantly in resolution and aspect ratio across hospitals and acquisitions. To standardize inputs while preserving aspect ratio, we employ letterbox resizing. Given a target canvas $(W_t, H_t) = (1280, 872)$, we compute a scaling factor $s = \min(W_t/W, H_t/H)$ and pad symmetrically. The NT bounding box coordinates are remapped via the affine transformation:
\[
x' = s \cdot x + \Delta x, \quad y' = s \cdot y + \Delta y
\]
where $(\Delta x, \Delta y)$ are the padding offsets. This ensures the ROI $\Omega$ is spatially consistent relative to the anatomical content.

\subsection{ROI-Aware Reconstruction Model}
\label{sec:model}

We adopt a Convolutional Autoencoder (CAE)~\cite{masci2011stacked} as the backbone for representation learning. The architecture is designed to compress the high-resolution input into a compact vector that forces the model to prioritize salient anatomical features.

\begin{figure}[t]
  \centering
  \includegraphics[width=\columnwidth]{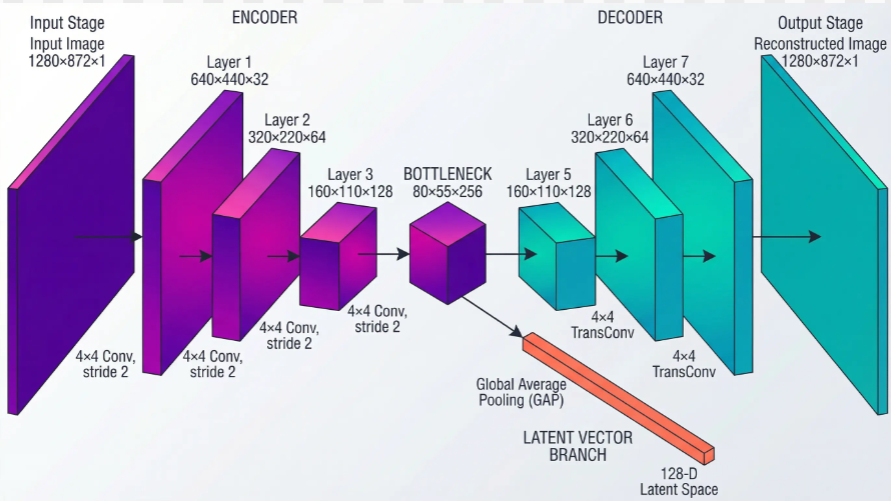}
  \caption{The proposed CAE. The encoder maps the input to a 128-D latent vector $z$ via global average pooling and a linear head. This compact $z$ serves as the shared representation for reconstruction and downstream tasks.}
  \label{fig:cae_architecture}
\end{figure}

\textbf{Architecture.} As shown in Figure~\ref{fig:cae_architecture}, the encoder $f_\theta$ processes the input $x$ into a bottleneck feature map $z_{\text{map}} \in \mathbb{R}^{256 \times 55 \times 80}$. To obtain a compact, task-agnostic representation, we apply Global Average Pooling (GAP)~\cite{lin2013network} followed by a linear projection to a 128-dimensional latent vector $z$:
\[
z = W_{\text{proj}} \cdot \text{GAP}(f_\theta(x)) + b, \quad z \in \mathbb{R}^{128}.
\]
The decoder $g_\phi$ then reconstructs the image $\hat{x} = g_\phi(z)$. This 128-D bottleneck is a critical design choice: it is small enough to act as an information bottleneck (filtering noise) but large enough to encode both anatomy and domain style.

\subsection{Two-Phase Optimization Strategy}
\label{sec:optimization}

To address the failure modes described in Section~\ref{sec:problem}, we propose a two-phase training strategy (Figure~\ref{fig:two_phase_overview}). Direct optimization of ROI losses from scratch often leads to unstable convergence; our approach first anchors the global structure before refining local details.

\textbf{Phase 1: Global Structural Warm-up.}
We first train the network to minimize the Multi-Scale Structural Similarity (MS-SSIM) loss on the entire image:
\[
\mathcal{L}_{\text{P1}} = 1 - \text{MS-SSIM}(x, \hat{x}).
\]
This phase ensures the latent space $z$ captures the global geometry and domain characteristics (e.g., speckle patterns) of the ultrasound image.

\textbf{Phase 2: ROI-Focused Refinement.}
In the second phase, we introduce constraints to sharpen the anatomy within the NT box $\Omega$. The objective function becomes a weighted sum of three terms:
\[
\mathcal{L}_{\text{P2}} = \lambda_{\text{glob}}\mathcal{L}_{\text{glob}} + \lambda_{\text{L1}}\mathcal{L}_{\text{ROI-L1}} + \lambda_{\text{edge}}\mathcal{L}_{\text{ROI-edge}}.
\]
\begin{itemize}
    \item \textbf{Global Anchor ($\mathcal{L}_{\text{glob}}$):} Maintains the structural fidelity learned in Phase 1.
    \item \textbf{ROI Intensity ($\mathcal{L}_{\text{ROI-L1}}$):} An L1 loss computed only over pixels $p \in \Omega$, enforcing pixel-level accuracy for the NT fluid.
    \item \textbf{ROI Edge Consistency ($\mathcal{L}_{\text{ROI-edge}}$):} To prevent boundary softening, we compute the Sobel gradient magnitude $M(\cdot)$~\cite{sobel19683x3} and minimize the difference within the ROI:
    \[
    \mathcal{L}_{\text{ROI-edge}} = \frac{1}{|\Omega|} \sum_{p \in \Omega} | \widetilde{M}(x)_p - \widetilde{M}(\hat{x})_p |.
    \]
    We use a normalized gradient magnitude $\widetilde{M}$ to make the loss robust to contrast variations between hospitals.
\end{itemize}

\textbf{Gradient-Based Weight Calibration.}
Balancing global and local losses is non-trivial. Instead of manual tuning, we employ a gradient calibration step at the start of Phase 2, inspired by GradNorm~\cite{chen2018gradnorm}. We compute the average gradient norms for each loss component on a calibration batch and set $\lambda_k$ inversely proportional to these norms. This ensures that the ROI refinement signals are not overwhelmed by the global reconstruction gradients during the critical fine-tuning steps.

\textbf{Implementation Details.}
We implement the framework in PyTorch. The model is trained using the Adam optimizer with a batch size of 8. Phase 1 uses a learning rate of $10^{-4}$ for 250 epochs, while Phase 2 uses a reduced learning rate of $10^{-5}$ to ensure stable refinement. All experiments are run on a single NVIDIA Tesla T4 GPU.

\subsection{Downstream Evaluation Framework}
\label{sec:downstream_framework}

We evaluate the utility of the learned representations $z$ and ROI-pooled features $\bar{z}_{\Omega}$ on three downstream tasks. In all cases, the encoder is \textbf{frozen} after Phase 2.

\textbf{Task 1: Provenance Classification.} We train a linear classifier on $z$ to predict the source hospital. This tests whether the representation encodes site-specific artifacts.
\textbf{Task 2: Quality Control (QC) Triage.} We predict measurement suitability using a feature vector $q(x) = [r_{\Omega},\, e_{\Omega},\, \|z\|_2]$, where $r_{\Omega}$ and $e_{\Omega}$ are ROI reconstruction residuals.
\textbf{Task 3: OOD Detection.} We score out-of-distribution samples using Mahalanobis distance on $z$ or KNN density, evaluating safety for deployment.

\section{Experiments}
\label{sec:experiments}

\subsection{Experimental Setup}
\label{sec:setup}
We evaluate our framework using a strict hospital-wise protocol. Hospitals 1 and 2 are used for training and validation, while Hospital 3 is held out exclusively for testing. This ensures that all reported results reflect true cross-site generalization.
For Phase 1, we train for 150 epochs with a learning rate of $10^{-4}$. For Phase 2, we refine for 100 epochs with a reduced learning rate of $10^{-5}$, using the gradient-based calibration described in Section~\ref{sec:optimization} to initialize loss weights. Unless otherwise stated, all experiments use Adam (batch size 8) with letterbox-resized inputs ($1280\times872$) and a 128-D latent code; results are averaged over 5 random seeds.

\subsection{Ablation Study: Loss Selection}
\label{sec:ablation}
To determine the optimal combination of objectives for Phase-2 refinement, we conducted a controlled ablation study on the Hospital-1/2 validation split. All runs were initialized from the same Phase-1 checkpoint and trained for a fixed horizon of 15 epochs.

\begin{table*}[h]
\centering
\caption{Phase-2 ablation study on the Hospital-1/2 \textbf{validation} split (15 epochs from the same Phase-1 checkpoint). ROI-$L_1$ best improves ROI intensity (ROI MAE), ROI-Sobel best improves boundary preservation (ROI Edge-MAE), and ROI-DINO does not help. These trends motivate the final Phase-2 objective (Global + ROI-$L_1$ + ROI-Sobel; no DINO) evaluated in the main results.}
\label{tab:ablation_val}
\begin{small}
\begin{sc}
\renewcommand{\arraystretch}{1.1}
\resizebox{\textwidth}{!}{
\begin{tabular}{lccccc}
\toprule
\textbf{Phase-2 added term(s)} & \textbf{PSNR} $\uparrow$ & \textbf{MS-SSIM} $\uparrow$ & \textbf{ROI MAE} $\downarrow$ & \textbf{ROI Edge-MAE} $\downarrow$ & \textbf{Notes} \\
\midrule
Baseline (P1 Init) & 37.37 & 0.9987 & 0.00566 & 0.02961 & -- \\
\midrule
+ ROI-$L_1$        & \textbf{37.52} & \textbf{0.9988} & \textbf{0.00518} & 0.02877 & Best intensity \\
+ ROI-Sobel (edge) & 37.47 & 0.9988 & 0.00527 & \textbf{0.02847} & Best edge \\
+ ROI-DINO         & 37.41 & 0.9987 & 0.00568 & 0.02982 & Degrades \\
+ ROI-$L_1$ + ROI-Sobel + ROI-DINO
                   & 37.43 & 0.9988 & 0.00522 & 0.02884 & Includes DINO \\
\bottomrule
\end{tabular}
}
\end{sc}
\end{small}
\end{table*}

Table~\ref{tab:ablation_val} summarizes the results. While the global-only baseline preserves overall structure, it exhibits limited ROI-specific fidelity. Adding the \textbf{ROI-$L_1$} term substantially reduces intensity error (ROI MAE $\downarrow$ 8.5\%), and the \textbf{ROI-Sobel edge} term most improves boundary preservation (ROI Edge-MAE $\downarrow$ 3.9\%). In contrast, the perceptual ROI-DINO loss does not yield improvements in this setting, likely because generic pre-trained features are not aligned with the pixel-precise requirements of NT measurement. Consequently, we select the combination \textbf{Global + ROI-$L_1$ + ROI-Sobel} (without DINO) for the final Phase-2 model. Loss selection is performed exclusively on the Hospital-1/2 validation split to avoid any exposure to the held-out test domain. For transparency, we additionally report the corresponding ablation outcomes on the held-out Hospital-3 test set in Appendix Table~\ref{tab:ablation_test_final}; these results are reported for reference only and are not used for model selection.

\subsection{Main Reconstruction Results}
\label{sec:main_results}
Table~\ref{tab:reconstruction_protocols} reports the final reconstruction performance under three site-held-out evaluation protocols. We utilize MS-SSIM \citep{wang2003multiscale} to assess global structural fidelity, alongside ROI-specific metrics to quantify anatomical preservation.

\begin{figure}[t]
    \centering
    \begin{subfigure}[t]{0.48\columnwidth}
        \centering
        \includegraphics[width=\linewidth]{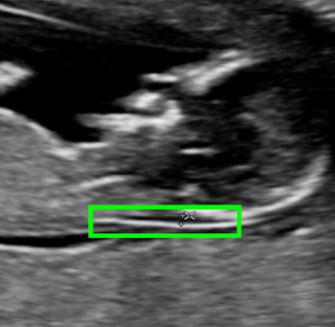}
        \caption{Original ultrasound image}
        \label{fig:qualitative_original}
    \end{subfigure}
    \hfill
    \begin{subfigure}[t]{0.48\columnwidth}
        \centering
        \includegraphics[width=\linewidth]{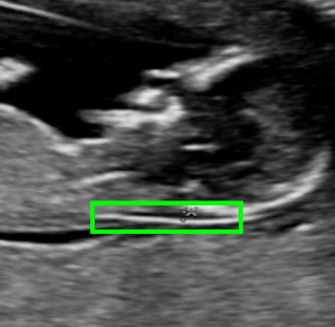}
        \caption{Phase-2 reconstruction}
        \label{fig:qualitative_reconstruction}
    \end{subfigure}

    \caption{Qualitative example on the held-out Hospital-3 test set. 
    Left: original ultrasound image. 
    Right: Phase-2 reconstruction. 
    The NT region (green box) shows improved boundary preservation, consistent with the ROI Edge-MAE reductions reported in Table~\ref{tab:reconstruction_protocols}.}
    \label{fig:qualitative_main}
\end{figure}

Across all domain-shift settings, Phase-2 consistently improves NT-focused metrics while maintaining or improving global image fidelity. Compared to Phase-1, Phase-2 reduces \textbf{ROI Edge-MAE} by approximately 4--12\% across held-out sites, while improving \textbf{ROI MS-SSIM} by up to 0.3--0.5 points. This indicates that our two-stage approach yields sharper and more structurally consistent NT reconstructions, even under significant domain shift.  A representative qualitative example is shown in Fig.~\ref{fig:qualitative_main}.

\textbf{Contextual Analysis.}
Improvements are most pronounced on NT-focused metrics, suggesting that Phase-2 selectively refines anatomically relevant regions rather than overfitting the full image distribution. This alignment with clinical ROI quality is critical; recent work by \citet{wang2025quality} demonstrated that intelligent quality assessment for NT measurement relies heavily on the precise delineation of the nasolabial sagittal plane.

\begin{table*}[t]
\centering
\caption{Phase-1 vs Phase-2 reconstruction performance under different train--test protocols 
(mean $\pm$ std over 5 random seeds).}
\label{tab:reconstruction_protocols}
\small
\setlength{\tabcolsep}{3.2pt} 
\renewcommand{\arraystretch}{1.10}
\begin{tabular}{llccccc}
\toprule
\textbf{Protocol} & \textbf{Phase} 
& \textbf{PSNR} $\uparrow$ 
& \textbf{MS-SSIM} $\uparrow$ 
& \textbf{ROI-MAE} $\downarrow$ 
& \textbf{ROI MS-SSIM} $\uparrow$ 
& \textbf{ROI Edge-MAE} $\downarrow$ \\
\midrule

H1+H2 $\rightarrow$ H3 
& Phase-1 
& 35.47 $\pm$ 0.56 
& 0.9969 $\pm$ 0.0006 
& 0.00929 $\pm$ 0.00053 
& 0.98645 $\pm$ 0.00165 
& 0.04062 $\pm$ 0.00634 \\
& Phase-2
& \textbf{35.59 $\pm$ 0.50} 
& \textbf{0.9970 $\pm$ 0.0005} 
& \textbf{0.00871 $\pm$ 0.00046} 
& \textbf{0.98716 $\pm$ 0.00142} 
& \textbf{0.03888 $\pm$ 0.00656} \\
\midrule

H1+H3 $\rightarrow$ H2
& Phase-1 
& 35.73 $\pm$ 0.05
& 0.9984 $\pm$ 0.0000
& 0.00695 $\pm$ 0.00017
& 0.9961 $\pm$ 0.0000
& 0.02766 $\pm$ 0.00069 \\
& Phase-2 
& \textbf{36.05 $\pm$ 0.21}
& \textbf{0.9985 $\pm$ 0.0001}
& \textbf{0.00609 $\pm$ 0.00035}
& \textbf{0.9965 $\pm$ 0.00028}
& \textbf{0.02423 $\pm$ 0.00140} \\
\midrule

H2+H3 $\rightarrow$ H1 
& Phase-1 
& 33.92 $\pm$ 0.58 
& 0.9969 $\pm$ 0.0007 
& 0.00634 $\pm$ 0.00048 
& 0.9941 $\pm$ 0.0010 
& 0.03549 $\pm$ 0.00402 \\
& Phase-2 
& \textbf{34.70 $\pm$ 0.55} 
& \textbf{0.9978 $\pm$ 0.0006} 
& \textbf{0.00528 $\pm$ 0.00045} 
& \textbf{0.99560 $\pm$ 0.0009} 
& \textbf{0.02920 $\pm$ 0.0038} \\
\bottomrule
\end{tabular}
\end{table*}

A complementary aggregate summary of Phase-2 gains under the standard dev split (Hosp-1/2 validation; Hosp-3 test), which is not one of the three site-held-out protocols in Table~\ref{tab:reconstruction_protocols}, is reported in Appendix Table~\ref{tab:phase2_delta_main}.

\subsection{Robustness Under Domain Shift}
\label{sec:robustness_domain_shift}

Beyond reconstruction fidelity, we evaluate the robustness and site-invariance of the learned representations using three complementary probes.

\textbf{Linear Probe for Site Predictability.}
We trained a linear classifier to predict the source hospital from latent vectors. On unseen hospitals, Phase-2 consistently yields lower maximum softmax confidence and higher entropy compared to Phase-1 (Table~\ref{tab:linear_probe_results}). For Test-H3, mean confidence decreases from $0.545$ to $0.528$, indicating reduced linear separability of hospital identity in the latent space.

\begin{table}[t]
\centering
\caption{Linear probe behavior on unseen hospitals (mean $\pm$ std).}
\label{tab:linear_probe_results}
\renewcommand{\arraystretch}{1.1}
\begin{tabular}{lccc}
\toprule
\textbf{Test Site} & \textbf{Phase} & \textbf{Confidence} & \textbf{Entropy} \\
\midrule
H1 & \textbf{Phase-1} & $\mathbf{0.751 \pm 0.118}$ & $\mathbf{0.523 \pm 0.132}$ \\
   & Phase-2 & $0.752 \pm 0.118$ & $0.521 \pm 0.132$ \\
\midrule
H2 & Phase-1 & $0.710 \pm 0.107$ & $0.574 \pm 0.099$ \\
   & Phase-2 & $\mathbf{0.710 \pm 0.107}$ & $\mathbf{0.575 \pm 0.099}$ \\
\midrule
H3 & Phase-1 & $0.556 \pm 0.035$ & $0.684 \pm 0.010$ \\
   & Phase-2 & $\mathbf{0.541 \pm 0.028}$ & $\mathbf{0.688 \pm 0.006}$ \\
\bottomrule
\end{tabular}
\end{table}

\begin{figure}[h]
\centering
\includegraphics[width=0.85\linewidth]{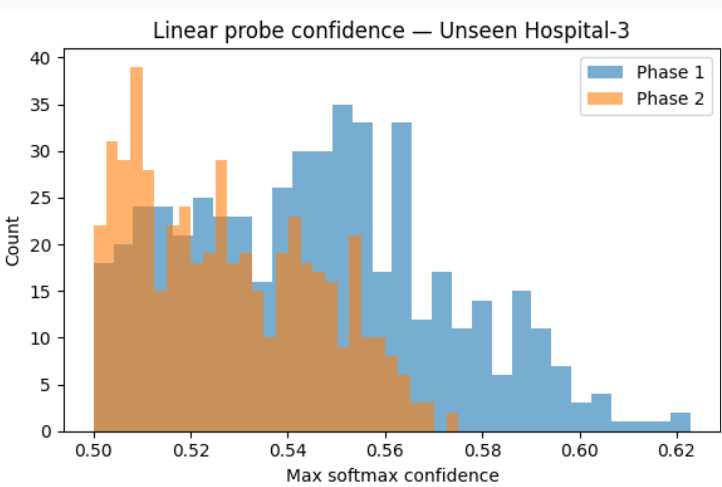}
\caption{Distribution of maximum softmax confidence for linear probe predictions on unseen hospitals. Phase-2 (orange) shifts left, indicating reduced site identifiability.}
\label{fig:linear_probe_confidence}
\end{figure}

\textbf{Latent-Space Out-of-Distribution Detection.}
We evaluated OOD detection using Mahalanobis and KNN distances. While global Mahalanobis scores remained stable near ceiling ($>0.99$), KNN-based detection improved consistently after Phase-2 (Table~\ref{tab:ood_results_singlecol}). For H1+H2$\rightarrow$H3, KNN AUROC increased from 0.9238 to 0.9251. This suggests Phase-2 enhances \emph{local} latent-space structure and neighborhood compactness.

\begin{table}[t]
\centering
\caption{Latent-space OOD detection performance (AUROC) under site-held-out protocols. Higher is better.}
\label{tab:ood_results_singlecol}
\renewcommand{\arraystretch}{1.05}
\resizebox{\columnwidth}{!}{
\begin{tabular}{lccc}
\toprule
\textbf{Protocol} & \textbf{Phase} & \textbf{Mahalanobis AUROC} $\uparrow$ & \textbf{KNN AUROC} $\uparrow$ \\
\midrule
H2 + H3 $\rightarrow$ H1 
    & Phase-1 & 0.8681 $\pm$ 0.0005 & 0.7152 $\pm$ 0.0013 \\
    & Phase-2 & \textbf{0.8684 $\pm$ 0.0003} & \textbf{0.7182 $\pm$ 0.0011} \\
\midrule
H1 + H3 $\rightarrow$ H2 
    & Phase-1 & 0.7400 $\pm$ 0.0004 & 0.6382 $\pm$ 0.0520 \\
    & Phase-2 & \textbf{0.7488 $\pm$ 0.0035} & \textbf{0.6421 $\pm$ 0.0547} \\
\midrule
H1 + H2 $\rightarrow$ H3 
    & Phase-1 & 0.9953 $\pm$ 0.0013 & 0.9317 $\pm$ 0.0079 \\
    &Phase-2 & \textbf{0.9956 $\pm$ 0.0014} & \textbf{0.9312 $\pm$ 0.0061} \\

\bottomrule
\end{tabular}
}
\end{table}

\begin{figure}[h]
    \centering
    \begin{subfigure}{0.48\columnwidth}
        \centering
        \includegraphics[width=\linewidth]{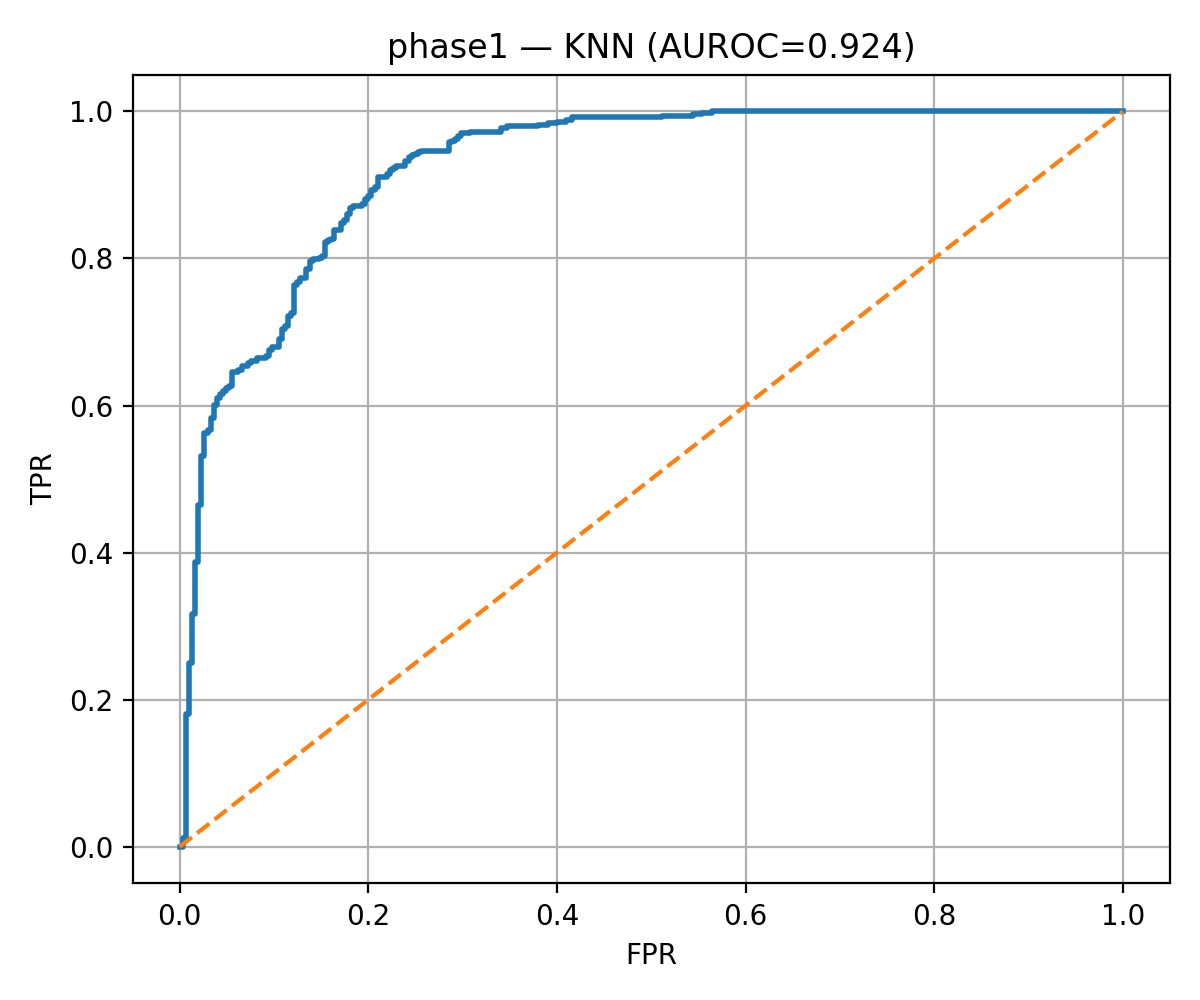}
        \caption{Phase-1 (AUROC = 0.924)}
    \end{subfigure}
    \hfill
    \begin{subfigure}{0.48\columnwidth}
        \centering
        \includegraphics[width=\linewidth]{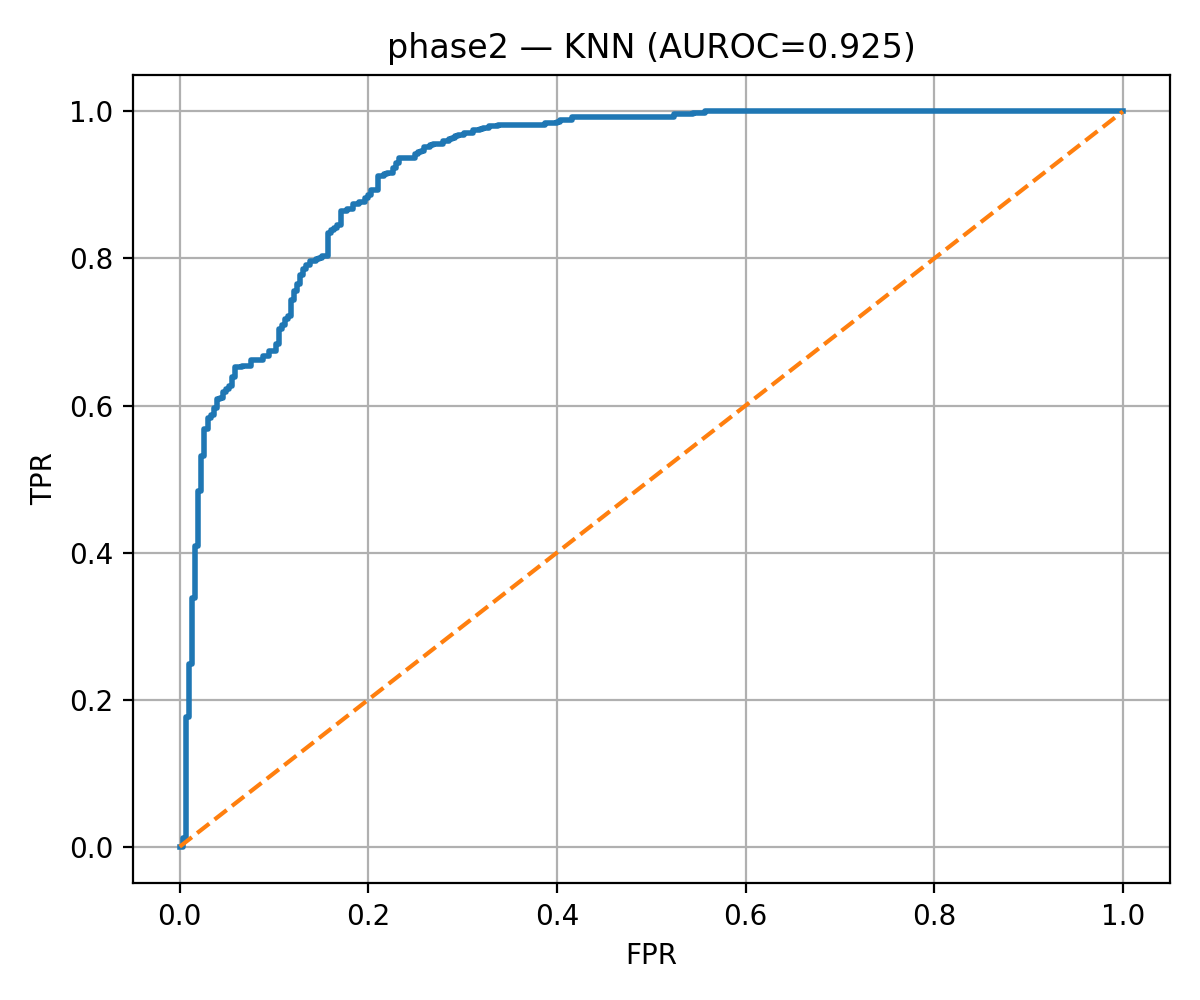}
        \caption{Phase-2 (AUROC = 0.925)}
    \end{subfigure}
    \caption{Latent-space OOD detection using KNN distance.}
    \label{fig:knn_ood_comparison}
\end{figure}

\textbf{Latent Quality-Control (QC) Probe.}
A linear probe trained to predict ROI edge error from the latent vector showed reduced $R^2$ on held-out sites after Phase-2 (Table~\ref{tab:qc_probe_hospitals}). This indicates weaker coupling between latent variables and local edge degradation, suggesting a more invariant latent space.

\begin{table}[t]
\centering
\caption{QC probe results on held-out test hospitals. Lower values indicate reduced leakage of reconstruction artifacts into the latent space. Results for Test H2 are reported once available.}
\label{tab:qc_probe_hospitals}
\renewcommand{\arraystretch}{1.1}
\resizebox{\columnwidth}{!}{
\begin{tabular}{lcccccc}
\toprule
\multirow{2}{*}{Model} 
& \multicolumn{2}{c}{Test H1} 
& \multicolumn{2}{c}{Test H2} 
& \multicolumn{2}{c}{Test H3} \\
\cmidrule(lr){2-3} \cmidrule(lr){4-5} \cmidrule(lr){6-7}
& $R^2$ & Spearman $\rho$ 
& $R^2$ & Spearman $\rho$ 
& $R^2$ & Spearman $\rho$ \\
\midrule
Phase-1 & 0.177 & 0.321 & 0.175 & 0.493 & 0.053 & 0.298 \\
Phase-2 & 0.235 & 0.317 & 0.153 & 0.407 & 0.074 & 0.358 \\

\bottomrule
\end{tabular}
}
\end{table}

\begin{figure}[h]
    \centering
        \includegraphics[height=0.14\textheight,keepaspectratio]{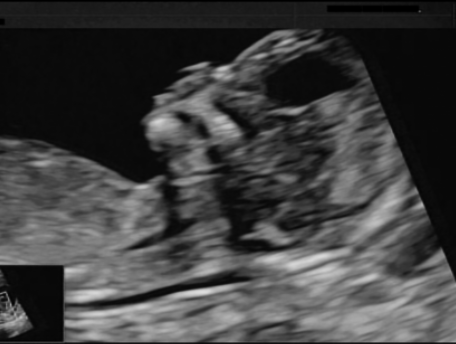}
        \includegraphics[height=0.14\textheight,keepaspectratio]{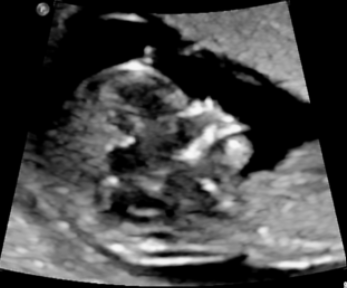}
    \caption{Latent space generation using the Phase-2 model. Perturbation over the latent representations produces anatomically coherent images, indicating a well-structured and stable latent manifold.}
    \label{fig:latent_interpolation}
\end{figure}

\textbf{Latent space interpolation.}
Figure~\ref{fig:latent_interpolation} visualizes linear interpolation between latent representations
extracted from real test images using the Phase-2 model.
The smooth and anatomically consistent transition between the endpoints confirms that Phase-2
learns a structured and continuous latent manifold without mode collapse.

\textbf{Latent-space structure under hospital shift (qualitative).}
To complement the quantitative provenance and OOD results, we visualize the frozen latent codes for the H1+H2 $\rightarrow$ H3 protocol using PCA and UMAP (Appendix~\cref{fig:latent_shift_h3}). The projections suggest that Phase-2 refinement preserves the global manifold geometry while leaving residual site-dependent structure observable, consistent with the downstream probe behavior under domain shift.
\section{Discussion}
\label{sec:discussion}

This work targets a common failure mode in clinical ultrasound learning: models can achieve excellent whole-image reconstruction while subtly degrading the small structures that drive measurements. In NT screening, such degradations (boundary softening, pixel-level shifts, and membrane attenuation) directly affect caliper placement yet can be weakly penalized by global metrics. Across strict hospital-wise splits, our results support the central hypothesis that incorporating ROI-aware constraints into representation learning is necessary to preserve measurement-critical anatomy under real cross-hospital domain shift.

\paragraph{Why the two-phase strategy improves generalization (not just sharpness).}
A key design choice is staging the optimization. Directly enforcing ROI constraints from scratch can destabilize training because ROI objectives emphasize high-frequency details and can dominate early gradients, leading to brittle solutions. Phase~1 provides a stable structural anchor via global MS-SSIM, encouraging the latent code to encode global geometry and coarse appearance in a scanner-agnostic way. Phase~2 then injects measurement-specific supervision only within $\Omega$, explicitly targeting NT failure modes while retaining the global anchor. Empirically, this staged refinement improves ROI fidelity while maintaining (and slightly improving) global fidelity across protocols (Table~\ref{tab:reconstruction_protocols}), indicating that ROI refinement is not merely a local sharpening trick but a controlled way to bias the representation toward transferable anatomy.

\paragraph{Gradient calibration as a transfer- and reproducibility-oriented mechanism.}
Multi-center ultrasound introduces not only visual domain shift but also loss-scale drift due to differences in gain, dynamic range compression, and vendor post-processing. In this setting, fixed hand-tuned loss weights can become site-dependent. Our gradient-based calibration initializes $\lambda_k$ from per-term gradient magnitudes at the start of Phase~2, enforcing comparable influence across objectives without manual tuning. Practically, this reduces sensitivity to arbitrary numeric scales and improves reproducibility across hospital-wise splits, which aligns with the paper’s emphasis on generalization under deployment-like evaluation.

\paragraph{ROI constraints: what transfers and what is domain-sensitive.}
The ablation study clarifies which ROI signals are robust under domain shift. ROI-$L_1$ yields consistent gains in ROI intensity and boundary-related metrics on both validation and the held-out hospital (Appendix Table~\ref{tab:ablation_test_final}), suggesting it captures a domain-robust notion of preserving NT appearance. By contrast, the Sobel-based edge term is effective for sharpening boundaries on seen domains but is more sensitive to cross-site speckle statistics and scanner edge enhancement; on the unseen site it can sometimes amplify site-specific high-frequency texture. This motivates our use of normalized gradients and careful balancing, and highlights a broader insight for ultrasound: ``edges'' are not purely anatomical—they are entangled with acquisition physics—so edge supervision should be applied conservatively when generalization is the primary objective.

\paragraph{Downstream probes as additional evidence of representation generalization.}
A core contribution of the paper is treating the encoder as a reusable representation rather than optimizing reconstruction alone. We therefore evaluate the frozen encoder using downstream probes designed to stress-test cross-site behavior. First, provenance classification becomes less confident on the unseen hospital after Phase~2 (max-softmax 0.556$\rightarrow$0.541; entropy 0.684$\rightarrow$0.688; Table~\ref{tab:linear_probe_results}), indicating reduced leakage of site-specific artifacts into the latent code. Second, OOD detection remains strong across site-held-out protocols: Mahalanobis AUROC stays near ceiling in two splits (e.g., 0.9953$\rightarrow$0.9956 for H1+H2$\rightarrow$H3) and improves in the more challenging split (0.7400$\rightarrow$0.7488 for H1+H3$\rightarrow$H2), with modest corresponding KNN gains (0.6382$\rightarrow$0.6421; Table~\ref{tab:ood_results_singlecol}). Third, the QC triage signal derived from ROI residuals remains informative under shift (e.g., on H3 Spearman 0.298$\rightarrow$0.358; Table~\ref{tab:qc_probe_hospitals}), supporting the view that the representation can be paired with monitoring signals when deployed out-of-distribution. Collectively, these probes provide a multi-angle ``proof-by-transfer'': gains are not limited to a single metric or a single hospital, and the representation remains useful for safety- and workflow-relevant tasks on unseen data.

\paragraph{Limitations and future work toward broader anatomical and clinical generalization.}
This study focuses on NT-view images and uses bounding boxes to define $\Omega$, which is a practical form of weak supervision but still requires annotation and may omit clinically relevant context (e.g., subtle plane-selection cues). Our evaluation includes three hospitals with one strictly held out; broader multi-vendor and multi-country studies are needed to further stress-test generalization. Methodologically, edge supervision remains the most domain-sensitive component, motivating future work on ROI-adaptive or noise-aware edge constraints (e.g., confidence-weighted gradients or SNR-aware weighting) and tighter coupling to view-quality criteria so that refinement is strongest when the acquisition is suitable for measurement. Importantly, the framework itself is ROI-agnostic: by redefining $\Omega$ and selecting measurement-aligned constraints, the same approach can be adopted for other fetal biometry targets such as CRL and NB, where boundary fidelity similarly determines measurement reliability. More broadly, ROI-aware refinement can extend to other medical imaging problems where small structures dominate clinical decisions, providing a general template for anatomy-preserving representation learning under domain shift.

\section{Conclusion}
\label{sec:conclusion}

We introduced a two-phase, ROI-aware representation learning framework for measurement-critical fetal ultrasound and demonstrated it on first-trimester NT screening under strict hospital-wise generalization. The model learns a compact 128-D latent representation anchored by global MS-SSIM, then refines the NT ROI using intensity and normalized edge constraints, with gradient-based calibration to stabilize multi-objective optimization without manual tuning. Under a fully held-out hospital protocol, ROI refinement improves measurement-relevant fidelity (ROI Edge-MAE reductions of 11.10\% on source hospitals and 4.90\% on the unseen hospital) while also improving global quality (e.g., +0.29\,dB PSNR on held-out test). Frozen-latent downstream probes further support cross-site generalization by showing reduced site identifiability on the unseen hospital and consistently high OOD-detection performance across protocols. Because the approach only requires an ROI definition and measurement-aligned constraints, it provides a general template that can be adopted for other fetal biometry targets such as CRL and nasal bone (NB), and more broadly for medical imaging problems where small anatomical regions dominate clinical decisions. Future work will focus on multi-ROI training, ROI acquisition from automatic detectors, and broader multi-vendor prospective validation.

\section*{Impact Statement}
\label{sec:broader_impact}

This work aims to improve the reliability of measurement-critical ultrasound AI under real clinical variation. In multi-center practice, scanners, acquisition protocols, and operator technique can shift image appearance in ways that cause models trained with global objectives to fail silently, especially when the clinically relevant anatomy occupies a small fraction of the image. By making the ROI explicit in the representation objective and by evaluating with a fully held-out hospital, we emphasize generalization and deployment realism as primary design goals rather than post-hoc considerations.

A potential positive impact is more robust automation for first-trimester screening workflows, where preserving NT anatomy can support downstream quality control, triage, and safer model operation in new hospitals. Importantly, the framework is anatomy-agnostic: the same ROI-aware refinement principle can be adopted for other fetal biometry targets such as crown--rump length (CRL) and nasal bone (NB), where boundary fidelity similarly determines measurement reliability. More broadly, redefining the ROI and constraints makes the approach applicable to other ultrasound measurements and to other medical imaging settings where small structures dominate clinical decisions.

\textbf{Risks remain.} Reconstruction-based models can introduce subtle artifacts or boundary shifts that might bias measurements if relied upon without safeguards, and generalization can degrade under acquisition regimes outside the training distribution (e.g., unseen vendors or handheld devices). To mitigate these risks, we advocate coupling deployment with explicit QC/triage and OOD monitoring, expanding evaluation to diverse multi-vendor cohorts, and performing prospective clinical validation before clinical use. Future work on adaptive, noise-aware ROI constraints and multi-ROI refinement is also intended to improve robustness while preserving the central focus on safe generalization.

\section{Acknowledgments} 
This publication was funded by the PPM7th Cycle grant (PPM 07-0409-240041, AMAL-For-Qatar) from the Qatar Research Development and Innovation Council (QRDI), a member of the Qatar Foundation. The findings herein reflect the work and are solely the responsibility, of the authors.

\bibliography{final}
\bibliographystyle{icml2026}

\newpage
\appendix
 \onecolumn

\section*{Appendix}
\section{Extended Literature Discussion}
\label{app:literature}

Our work relates to three lines of research: ultrasound image enhancement and reconstruction, ROI-aware and boundary-preserving learning in medical imaging, and loss weighting for multi-objective optimization. The central theme connecting these areas is that \emph{global} image quality improvements do not necessarily translate to \emph{measurement-relevant} fidelity when clinically important anatomy occupies a small region of the image, as in NT assessment.

\subsection{Ultrasound Image Enhancement and Reconstruction}
Ultrasound enhancement differs from many restoration problems because the image formation process produces speckle---a multiplicative, spatially correlated phenomenon whose texture can resemble anatomy and whose suppression can inadvertently remove subtle boundaries \cite{hu2024,michailovich2006}. Classical despeckling and filtering methods therefore face a persistent trade-off between noise reduction and boundary preservation.

Deep learning has substantially advanced medical image reconstruction through encoder--decoder paradigms. U-Net \cite{ronneberger2015} and its extensions, including 3D variants \cite{cicek2016}, nested skip designs \cite{zhou2018}, and multi-resolution feature aggregation \cite{ibtehaz2020}, have provided strong backbones for reconstruction and enhancement. Residual learning approaches such as DnCNN \cite{zhang2017} further improved denoising by explicitly predicting the corruption component rather than regressing the clean image directly.

Generative models have been widely used for image-to-image translation and enhancement, especially when paired supervision is limited. Conditional GANs \cite{isola2017} enable supervised translation, while CycleGAN \cite{zhu2017} extends this to unpaired settings and has been used for medical domain translation and adaptation \cite{sandfort2019}. For ultrasound, CycleGAN-style enhancement has been combined with perceptual objectives to improve visual quality under unpaired data \cite{athreya2024cyclegan}. More recently, diffusion probabilistic models have emerged as powerful priors for restoration and reconstruction across modalities \cite{kazerouni2024}, including high-fidelity reconstruction settings such as accelerated MRI \cite{chung2024}. These generative approaches are attractive when the desired output distribution is complex; however, in measurement-oriented clinical workflows, visual plausibility alone can be insufficient if fine boundaries in small anatomical regions are altered.

\paragraph{Fetal reconstruction context (3D from routine acquisitions).}
A related body of work in fetal imaging tackles reconstruction under severe motion and variable acquisition coverage. In fetal MRI, automated slice-to-volume registration pipelines enable robust reconstruction from motion-corrupted stacks with reduced manual input \cite{UUS2022102484}. In fetal ultrasound, several methods aim to recover 3D anatomical context from routine 2D acquisitions: conditional generative models reconstruct fetal anatomy (e.g., skull) from standard 2D planes \cite{tan2024}, implicit neural representations reconstruct volumes from non-tracked freehand sweeps \cite{YEUNG2024103147}, and patch-based fusion improves robustness to inconsistencies such as motion artifacts without explicitly modeling them \cite{Gomez2019}. While these works address volumetric reconstruction rather than first-trimester NT screening, they reinforce a key point relevant to our setting: reconstruction quality should be assessed by whether clinically relevant structures remain stable under motion/noise and device/operator variability, not only by global appearance.

\paragraph{Reconstruction objectives beyond pixel losses.}
Loss functions strongly influence whether fine detail is preserved or smoothed. SSIM \cite{wang2004} evaluates structural similarity via local luminance, contrast, and structure statistics and is often better aligned with perceived fidelity than MSE. Perceptual losses computed in the feature space of pretrained networks can preserve high-frequency content that pixel-wise losses tend to suppress \cite{johnson2016}, and have been adapted for medical enhancement tasks \cite{ma2022}. In our setting, these observations motivate anchoring training with a global structural objective while adding ROI-local constraints that directly target NT boundary and intensity fidelity.

\paragraph{Fetal ultrasound and NT-oriented automation.}
Fetal ultrasound introduces additional challenges due to fetal motion, acquisition-plane sensitivity, and operator and device variability. Prior work has addressed fetal imaging artifacts (e.g., shadows) with learning-based methods \cite{meng2019} and has explored super-resolution for fetal ultrasound quality enhancement \cite{boumeridja2025}. Concurrently, NT-specific automation has progressed toward clinical-grade assessment and measurement \cite{kasera2024,hur2024,li2024nt,wang2025quality}. Practical studies also highlight the importance of reproducibility and training for reliable NT measurement, including work using 3D-printed fetal phantoms to assess measurement accuracy across examiners \cite{Recker2024}. While these efforts demonstrate the feasibility of AI-driven NT analysis, enhancement and reconstruction components are still commonly optimized with \emph{global} objectives, leaving open the question of how to explicitly preserve measurement-critical NT boundaries under realistic cross-site variability.

\subsection{ROI-Aware and Attention-Based Learning}
Because clinically relevant information may be concentrated in a small portion of an image, ROI-aware learning has been extensively studied in medical imaging. A prominent direction is attention mechanisms that suppress irrelevant background while emphasizing salient regions. Attention U-Net \cite{oktay2018} introduced attention gates that improve focus on target anatomy, and channel attention via squeeze-and-excitation modules further improves feature recalibration \cite{rundo2019}. Transformer-based architectures extend this idea by modeling longer-range dependencies; representative examples include TransUNet \cite{chen2021}, TransFuse \cite{zhang2021}, and transformer-based skip-connection designs \cite{wang2022}. Recent work continues to refine transformer-based segmentation with improved fusion and attention strategies \cite{Liyao2024sstrans,zhang2025feswu,golkhatmi2025swin,chen2025medfusenet}, and ViT-based methods have also been applied to ROI detection \cite{Hossain2023}. These approaches generally learn ROIs implicitly, whereas our setting provides an explicit NT bounding box that can be leveraged for direct ROI-supervised reconstruction.

\paragraph{Region-based losses and small-structure emphasis.}
Loss functions developed for segmentation highlight how supervision can be shaped toward small or imbalanced targets. Dice loss \cite{milletari2016} and its generalized variant \cite{sudre2017} mitigate class imbalance, while Tversky loss \cite{salehi2017} allows explicit control over false positives versus false negatives. Focal loss \cite{lin2017} down-weights easy examples to emphasize hard cases, and blob-level formulations address instance-level imbalance \cite{kofler2023}. While our task is reconstruction rather than segmentation, the underlying principle applies: when the clinically important region is small, objectives should be constructed to ensure that optimization pressure is not dominated by background pixels.

\paragraph{Boundary-aware supervision.}
Boundary preservation is particularly important for measurement tasks where small spatial shifts can impact clinical metrics. Hausdorff-distance-inspired objectives \cite{karimi2020} and distance-based boundary losses \cite{kervadec2021} explicitly target contour alignment, and broader evaluations show that combining complementary region and boundary terms often yields the best performance for small structures \cite{ma2021}. Recent methods continue to refine boundary supervision and differentiability, including cbDice \cite{pan2024cbdice} and sub-differentiable Hausdorff losses \cite{visalakshi2025sdhl}, as well as approaches for ambiguous boundaries \cite{wang2024condseg} and architectures that explicitly integrate boundary extraction \cite{chen2024bgdnet}. These works primarily address segmentation; our contribution transfers boundary-aware principles to \emph{reconstruction} by enforcing Sobel-based edge consistency inside a known NT ROI, enabling explicit penalties for boundary softening and local edge drift without dense boundary annotations.

\subsection{Multi-Task Learning and Loss Weighting}
Training with multiple objectives is standard in medical imaging pipelines, but heterogeneous loss scales and differing convergence rates can make optimization brittle. Na\"ive fixed-weight summation may cause one term to dominate, leaving other objectives under-optimized.

\paragraph{Uncertainty- and gradient-based balancing.}
Uncertainty-based weighting \cite{kendall2018} learns task weights via homoscedastic uncertainty and has been widely adopted, though it can be sensitive to initialization. GradNorm \cite{chen2018gradnorm} directly addresses scale mismatch by adjusting weights to equalize gradient norms across tasks, promoting balanced learning progress. Related ideas include trajectory-based weighting such as SoftAdapt \cite{heydari2019}, and multi-objective formulations that seek Pareto-optimal trade-offs \cite{sener2018}. Broader surveys summarize architectures and optimization strategies for dense prediction and multi-task learning \cite{jadon2020,amyar2020,vandenhende2022}.

\paragraph{Recent advances.}
Recent work has proposed analytically grounded uncertainty weighting with softmax normalization \cite{ Kirchdorfer2025}, bilevel loss balancing with theoretical guarantees \cite{Peiyao2025bilb}, scalable balancing for large shared-backbone systems \cite{liu2024multibalance}, efficient task affinity estimation \cite{Dongyue2024gradtag}, and alternative multi-objective descent perspectives \cite{Jacobian2024jacobian}. Region-specific weighting has also been explored in medical segmentation to adapt loss emphasis spatially \cite{Chen2024adaptive}, and hybrid uncertainty--hardness approaches have been validated across architectures \cite{Shasha2022ahumulti,zhu2025uncertainty}. A recent survey emphasizes that effective multi-task learning requires both appropriate model design and careful optimization \cite{ Yu2025Multitask}.

\paragraph{Positioning of our approach.}
Our strategy is inspired by gradient-based balancing \cite{chen2018gradnorm} but targets a different objective: rather than dynamically adapting weights throughout training, we use per-term gradient magnitudes to \emph{initialize} loss weights so that each term has comparable influence at the onset of ROI refinement. This yields a reproducible starting point that reduces sensitivity to loss scale differences without adding continuous adaptation complexity. Combined with a two-stage protocol---global warm-up followed by ROI-aware refinement---this design directly addresses the common failure mode in ultrasound reconstruction where global metrics improve while measurement-critical ROI boundaries degrade, ensuring the learned latent representation remains robust for downstream clinical tasks. 

\section{Implementation Details}
\label{app:implementation}

\subsection{Architecture Specification}
We employ a CAE with a symmetric encoder-decoder structure. The encoder maps the input image $x \in \mathbb{R}^{1 \times 1280 \times 872}$ to a compact latent vector $z \in \mathbb{R}^{128}$.

\textbf{Encoder.} The encoder consists of four convolutional blocks. Each block performs a convolution with kernel size $4 \times 4$, stride 2, and padding 1, followed by a LeakyReLU activation ($\alpha=0.1$). This results in a downsampling factor of 16. The final feature map is passed through a bottleneck convolution ($1 \times 1$ kernel) to fix the channel dimension to 256, followed by Global Average Pooling (GAP) and a linear projection to the 128-dimensional latent space.

\textbf{Decoder.} The decoder mirrors the encoder using transposed convolutions (kernel $4 \times 4$, stride 2, padding 1) to upsample the feature map back to the original resolution. The final layer uses a Sigmoid activation to bound the output in $[0, 1]$.

\begin{table*}[h]
    \centering
    \caption{Detailed architecture of the proposed CAE. Input size: $1 \times 1280 \times 872$.}
    \label{tab:arch_details}
    \small
    \begin{tabular}{lcccc}
    \toprule
    \textbf{Stage} & \textbf{Layer Type} & \textbf{Kernel / Stride} & \textbf{Channels (In $\to$ Out)} & \textbf{Output Size} \\
    \midrule
    \textbf{Encoder} & Conv2d + LeakyReLU & $4\times4$ / 2 & $1 \to 32$ & $640 \times 436$ \\
                     & Conv2d + LeakyReLU & $4\times4$ / 2 & $32 \to 64$ & $320 \times 218$ \\
                     & Conv2d + LeakyReLU & $4\times4$ / 2 & $64 \to 128$ & $160 \times 109$ \\
                     & Conv2d + LeakyReLU & $4\times4$ / 2 & $128 \to 256$ & $80 \times 55$ \\
                     & Conv2d (Bottleneck) & $1\times1$ / 1 & $256 \to 256$ & $80 \times 55$ \\
    \midrule
    \textbf{Latent}  & AdaptiveAvgPool2d & - & $256$ & $1 \times 1$ \\
                     & Flatten + Linear  & - & $256 \to 128$ & $128$ \\
    \midrule
    \textbf{Decoder} & Linear + Unflatten & - & $128 \to 256 \times 80 \times 55$ & $80 \times 55$ \\
                     & ConvTranspose2d + LeakyReLU & $4\times4$ / 2 & $256 \to 128$ & $160 \times 109$ \\
                     & ConvTranspose2d + LeakyReLU & $4\times4$ / 2 & $128 \to 64$ & $320 \times 218$ \\
                     & ConvTranspose2d + LeakyReLU & $4\times4$ / 2 & $64 \to 32$ & $640 \times 436$ \\
                     & ConvTranspose2d + Sigmoid   & $4\times4$ / 2 & $32 \to 1$ & $1280 \times 872$ \\
    \bottomrule
    \end{tabular}
\end{table*}

\subsection{Training Hyperparameters}
All experiments were conducted using PyTorch on a single NVIDIA Tesla T4 GPU (Google Colab environment). We report results averaged over 5 independent runs with fixed seeds: $\{1000, 1001, 1002, 1003, 1004\}$.

\begin{table}[h]
    \centering
    \caption{Hyperparameters for Phase 1 (Global Warm-up) and Phase 2 (ROI Refinement).}
    \label{tab:hyperparams}
    \small
    \begin{tabular}{lcc}
    \toprule
    \textbf{Parameter} & \textbf{Phase 1} & \textbf{Phase 2} \\
    \midrule
    Optimizer & Adam & Adam \\
    Learning Rate & $1 \times 10^{-4}$ & $1 \times 10^{-5}$ \\
    Batch Size & 8 & 8 \\
    Max Epochs & 250 & 250 \\
    Early Stopping Patience & 5 epochs & 7 epochs \\
    Min Delta (Early Stop) & $2 \times 10^{-5}$ & $5 \times 10^{-5}$ \\
    Loss Function & $1 - \text{MS-SSIM}$ & Weighted Sum  \\
    \bottomrule
    \end{tabular}
\end{table}

\section{Detailed Experimental Protocol}
\label{app:protocol}

\subsection{Preprocessing: Letterbox and ROI Remapping}
To handle the variable aspect ratios and resolutions of clinical ultrasound images, we employ a ``Letterbox'' preprocessing strategy that preserves the original aspect ratio without distortion.

\textbf{Letterbox Resizing.} Given a target canvas of $W_t \times H_t = 1280 \times 872$, we compute a scaling factor $s = \min(W_t/W, H_t/H)$. The image is resized to $W' \times H'$ using bilinear interpolation. The resized image is then centered on the canvas, and the remaining areas are padded with zero (black).

\textbf{ROI Remapping.} The NT bounding box annotations must be transformed to match the resized coordinate system. For an original box $(x, y, w, h)$, the new coordinates $(x', y', x'_2, y'_2)$ are computed as:
\begin{align}
    x' &= s \cdot x + \text{offset}_x, \quad & y' &= s \cdot y + \text{offset}_y \\
    x'_2 &= s \cdot (x+w) + \text{offset}_x, \quad & y'_2 &= s \cdot (y+h) + \text{offset}_y
\end{align}
where $\text{offset}_x = (W_t - W') / 2$ and $\text{offset}_y = (H_t - H') / 2$. Boxes that fall outside the canvas or become degenerate after resizing are discarded.

\subsection{Downstream Validation Probes}
To assess the utility of the learned representations beyond reconstruction, we implement three probing tasks. In all cases, the CAE encoder is \textbf{frozen}, and only the probe heads are trained.

\paragraph{1. Linear Probe (Provenance Classification).}
We train a linear classifier to predict the source hospital from the latent vector $z$.
 
\begin{itemize} \item \textbf{Input:} Frozen latent vector $z \in \mathbb{R}^{128}$. \item \textbf{Model:} Single linear layer \texttt{nn.Linear(128, num\_classes)}. \item \textbf{Classes:} Hospital-1 and Hospital-2 (Seen domains). \item \textbf{Training:} 100 epochs, Adam optimizer, learning rate $1\times10^{-2}$, weight decay $1\times10^{-4}$. \item \textbf{Evaluation:} Accuracy and AUROC on the held-out Hospital-3 (Unseen domain) to analyze domain shift. 
\end{itemize}

\paragraph{2. OOD Detection (Mahalanobis \& KNN).}
We evaluate the ability to detect Out-of-Distribution (OOD) samples (Hospital-3) using the latent statistics of In-Distribution (ID) samples (Hospital-1 \& 2).
\begin{itemize}
    \item \textbf{Mahalanobis Distance:} We fit a multivariate Gaussian distribution ($\mu, \Sigma$) to the training set latents. The anomaly score for a test sample $z$ is the Mahalanobis distance: $d(z) = \sqrt{(z - \mu)^T \Sigma^{-1} (z - \mu)}$.
    \item \textbf{K-Nearest Neighbors (KNN):} We use a KNN detector with $K=10$ and Euclidean distance. The anomaly score is the average distance to the $K$ nearest training neighbors.
    \item \textbf{Metric:} AUROC separating ID (Hosp-1/2) from OOD (Hosp-3).
\end{itemize}

\paragraph{3. Quality Control (QC) Probe.}
We investigate if the latent space captures information about the reconstruction quality of the critical NT region.
\begin{itemize}
    \item \textbf{Target:} The ``ROI Edge Error'', defined as the Mean Absolute Error (MAE) between the normalized Sobel magnitude maps of the original and reconstructed NT regions.
    \item \textbf{Model:} Ridge Regression ($\alpha=1.0$).
    \item \textbf{Task:} Regress the ROI Edge Error directly from the latent vector $z$.
    \item \textbf{Evaluation:} $R^2$ score and Spearman rank correlation on the test set. A high correlation indicates that the latent code implicitly encodes the difficulty or quality of the anatomical reconstruction.
\end{itemize}

\begin{table*}[!t]
\centering
\caption{Phase-2 loss components evaluated on the held-out Hospital-3 test set (report-only; not used for loss selection)}

\label{tab:ablation_test_final}
\renewcommand{\arraystretch}{1.1}
\resizebox{0.8\textwidth}{!}{
\begin{tabular}{lccccc}
\toprule
\textbf{Phase-2 added term(s)} & \textbf{PSNR} $\uparrow$ & \textbf{MS-SSIM} $\uparrow$ & \textbf{ROI MAE} $\downarrow$ & \textbf{ROI Edge-MAE} $\downarrow$ & \textbf{Notes} \\
\midrule
Baseline (P1 Init) & 35.08 & 0.9965 & 0.00964 & 0.04320 & -- \\
\midrule
+ ROI-$L_1$        & 35.25 & \textbf{0.9966} & 0.00916 & \textbf{0.04256} & Best transfer (edge) \\
+ ROI-Sobel (edge) & \textbf{35.28} & \textbf{0.9966} & \textbf{0.00911} & 0.04439 & Best ROI MAE; edge degrades \\
+ ROI-DINO         & 35.13 & 0.9965 & 0.00963 & 0.04305 & No gain \\
+ ROI-$L_1$ + ROI-Sobel + ROI-DINO
                   & 35.22 & 0.9965 & 0.00920 & 0.04445 & Includes DINO; edge degrades \\
\bottomrule
\end{tabular}
}
\end{table*}

\section{Extended Experimental Results}
\label{app:extended_results}
\subsection{Extended Ablation Study (Generalization to Unseen Hospital)}
This section extends the Phase-2 validation ablation in Table~\ref{tab:ablation_val} by reporting the same fixed-horizon variants on the held-out Hospital-3 test set (Table~\ref{tab:ablation_test_final}), for report-only characterization under domain shift. These experiments were conducted using a dedicated ablation configuration with a fixed 15-epoch refinement horizon to isolate the contribution of each loss term; Hospital-3 is not used for model selection.

\textbf{Observation.} Under domain shift, \textbf{ROI-$L_1$} provides the most consistent improvement in boundary preservation (best ROI Edge-MAE). In contrast, the \textbf{ROI-Sobel edge} term can be sensitive to site-specific texture and noise, improving ROI MAE but degrading ROI Edge-MAE on the unseen hospital (edge amplification). The \textbf{ROI-DINO} term provides no meaningful gain. Overall, these results suggest ROI-$L_1$ transfers robustly, while edge supervision should be applied conservatively (e.g., via robust normalization and/or smaller weights) when targeting cross-hospital generalization.

\subsection{Aggregate Phase-2 Gains Under the Standard Dev Split}
To complement the protocol-wise results in Table~\ref{tab:reconstruction_protocols}, we report an aggregate summary of Phase-2 improvement over Phase-1 under the \emph{standard} development split used throughout model iteration: Hospital-1/2 for validation and held-out Hospital-3 for testing. Values are averaged over 5 random seeds. For MAE metrics, improvements are reported as relative percentage reduction ($\downarrow$); for MS-SSIM/PSNR, improvements are reported as absolute deltas ($\uparrow$). This summary is provided for convenience and should not be directly conflated with the three site-held-out protocols in Table~\ref{tab:reconstruction_protocols}.

\begin{table}[h]
\centering
\caption{Aggregate Phase-2 improvement over Phase-1 under the \textbf{standard dev split} (Hosp-1/2 validation; Hosp-3 test), averaged over 5 random seeds. MAE improvements are relative \% reductions ($\downarrow$); MS-SSIM/PSNR are absolute deltas ($\uparrow$).}
\label{tab:phase2_delta_main}
\small
\begin{sc}
\resizebox{0.8\textwidth}{!}{
\begin{tabular}{lcc}
\toprule
\textbf{Metric} & \textbf{Validation (H1/2) Improvement} & \textbf{Test (H3) Improvement} \\
\midrule
PSNR (dB)        & +0.27 dB  & +0.29 dB \\
MS-SSIM (full)   & +0.0002 & +0.0003 \\
\midrule
\textbf{ROI MAE}       & \textbf{-8.87\%} & \textbf{-6.43\%} \\
ROI MS-SSIM            & +0.0006 & +0.0013 \\
\textbf{ROI Edge-MAE}  & \textbf{-11.10\%} & \textbf{-4.90\%} \\
\bottomrule
\end{tabular}
}
\end{sc}
\end{table}

\subsection{Latent Space Visualizations Under Hospital Shift}
\label{app:latent_viz}

\begin{figure*}[t]
\centering
\begin{subfigure}{0.48\textwidth}
    \centering
    \includegraphics[width=\linewidth]{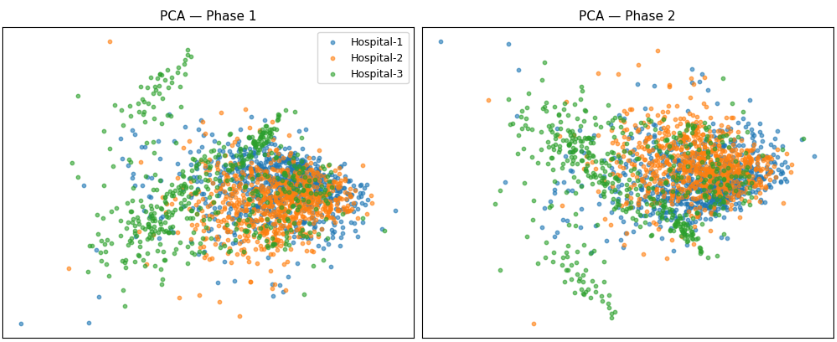}
    \caption{PCA (Phase-1 vs Phase-2)}
\end{subfigure}
\hfill
\begin{subfigure}{0.48\textwidth}
    \centering
    \includegraphics[width=\linewidth]{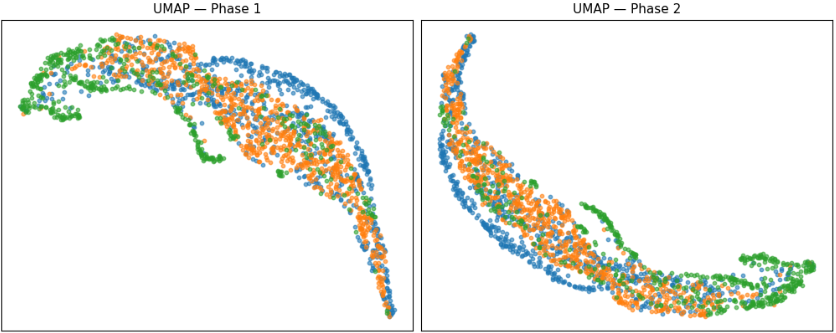}
    \caption{UMAP (Phase-1 vs Phase-2)}
\end{subfigure}
\caption{Latent space visualizations for the Hospital-1/2 $\rightarrow$ Hospital-3 split (frozen encoder), with points colored by hospital site. \textbf{(a)} PCA projections for Phase-1 (left) and Phase-2 (right) show strong overlap across hospitals and no major change in global latent geometry after refinement. \textbf{(b)} UMAP projections for Phase-1 (left) and Phase-2 (right) reveal site-dependent manifold structure, with held-out Hospital-3 samples occupying partially distinct regions, indicating observable domain shift. Phase-2 refinement preserves this structure without collapsing or artificially aligning domains.}
\label{fig:latent_shift_h3}
\end{figure*}

We provide qualitative latent-space visualizations for the Hospital-1/2 $\rightarrow$ Hospital-3 split to complement the quantitative probe results reported in the main paper. Specifically, \cref{fig:latent_shift_h3} projects frozen 128-D encoder representations using a linear method (PCA) and a nonlinear method (UMAP), with points colored by acquisition site. PCA shows substantial overlap among hospitals in both Phase-1 and Phase-2 and does not indicate a drastic re-organization of the global latent geometry after refinement. In contrast, UMAP reveals site-dependent manifold structure, with the held-out Hospital-3 samples tending to occupy partially distinct regions, indicating that domain shift remains observable in the learned representation. Importantly, Phase-2 refinement preserves this overall structure rather than collapsing or artificially aligning sites, which is consistent with our main findings: hospital information becomes less trivially predictable by simple probes, while residual shift remains exploitable for monitoring and OOD detection.

\end{document}